%% file: main.tex
\documentclass[runningheads]{llncs}
\usepackage[mobile]{eccv}

\usepackage{eccvabbrv}

\usepackage{colortbl}
\usepackage{amsmath,amsfonts,bm}
\usepackage{booktabs}       %
\usepackage{nicefrac}       %
\usepackage{microtype}      %
\usepackage{graphicx}
\usepackage{multirow}
\usepackage{caption}
\usepackage{subcaption}
\usepackage{physics}
\usepackage{rotating}
\usepackage{varwidth}
\usepackage{adjustbox}
\usepackage{arydshln}
\usepackage{microtype}      %
\usepackage{makecell}

\usepackage[misc]{ifsym}

\usepackage[accsupp]{axessibility}  

\input{configs}
\input{math_commands}

\renewcommand\paragraph[1]{\vspace{1.8mm}\noindent\textbf{#1}.}

\begin{document}

\title{Zero-Shot Image Feature Consensus\\with Deep Functional Maps}
\titlerunning{Zero-Shot Image Feature Consensus with Deep Functional Maps}

\author{%
    Xinle Cheng$^{1}$, Congyue Deng$^{2,\,\textrm{\Letter}}$, Adam Harley$^2$,\\
    Yixin Zhu$^{1,3,\,\textrm{\Letter}}$, Leonidas Guibas$^{2,\,\textrm{\Letter}}$\\
    \vspace{.4em}
    {\footnotesize $^{\textrm{\Letter}}$ congyue@stanford.edu, yixin.zhu@pku.edu.cn, guibas@stanford.edu
    }
}

\institute{
\vspace{-.3em}
\footnotesize $^1$ Institute for AI, Peking University, China \\
$^2$ Department of Computer Science, Stanford University, USA \\
$^3$ PKU-WUHAN Institute for Artificial Intelligence, China
}

\maketitle

\vspace{-1.5em}
\begin{abstract}
  Correspondences emerge from large-scale vision models trained for generative and discriminative tasks. This has been revealed and benchmarked by computing correspondence maps between pairs of images, using nearest neighbors on the feature grids. Existing work has attempted to improve the quality of these correspondence maps by carefully mixing features from different sources, such as by combining the features of different layers or networks. We point out that a better correspondence strategy is available, which directly imposes structure on the correspondence field: the functional map. Wielding this simple mathematical tool, we lift the correspondence problem from the pixel space to the function space and directly optimize for mappings that are globally coherent. We demonstrate that our technique yields correspondences that are not only smoother but also more accurate, with the possibility of better reflecting the knowledge embedded in the large-scale vision models that we are studying. Our approach sets a new state-of-the-art on various dense correspondence tasks. We also demonstrate our effectiveness in keypoint correspondence and affordance map transfer.
\end{abstract}
\vspace{-2em}

\begin{figure}
    \centering
    \includegraphics[width=\linewidth]{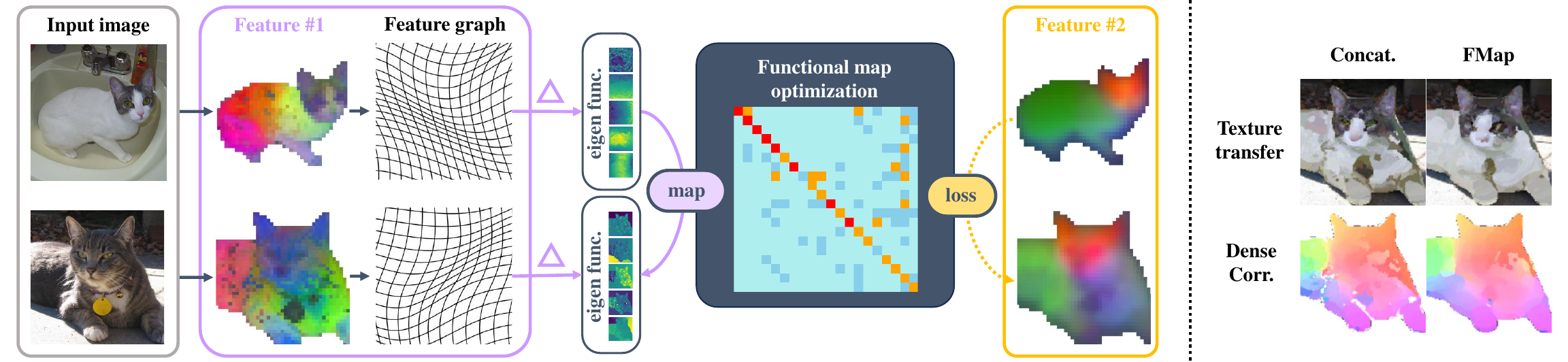}
    \vspace{-1.5em}
    \caption{\textbf{Overview.}
    \textbf{Left:} Given two sets of features, $E^M, E^N$, and $F^M, F^N$, we compute the Laplacian eigenfunction basis with $E^M, E^N$, and apply regularizations to the functional map optimization using $F^M, F^N$. This method optimizes a mapping in the spectral domain derived from one feature set to achieve a \textit{consensus} with the other set.
    \textbf{Right:} With a better understanding of the global image structure, our method produces smoother and more accurate correspondences in a zero-shot manner.}
    \label{fig:teaser}%
\end{figure}

\section{Introduction}
\label{sec:intro}
\vspace{-.8em}

Identifying image correspondence is a crucial task in mid-level computer vision. Recent advancements in large-scale vision models, trained for either generative~\cite{rombach2022high} or discriminative~\cite{caron2021emerging,oquab2023dinov2} tasks, possess emerged capabilities for dense correspondences~\cite{amir2021deep,hedlin2023unsupervised,tang2023emergent,zhang2023tale}. This learning is primarily facilitated by computing nearest neighbor matches between image patches with their feature similarities. Notably, the correspondences induced by these models can achieve comparable or even better performances compared to the methods explicitly designed for this purpose. However, a notable limitation arises: these models often struggle to retain the global structure of the correspondences. This can be attributed to the distortions and discontinuities in the nearest-neighbor search process.

While contemporary methods~\cite{zhang2023tale} have attempted to mitigate this problem by integrating features from different layers and networks, this approach only \textit{indirectly} confronts the fundamental issue---the lack of \textit{structure} in the correspondence maps. Fundamentally, point-wise correspondences are inherently susceptible to noise. Therefore, imposing a global structure on the correspondence maps is crucial for attaining high-quality correspondences without supervision

In this work, we leverage the \textit{functional maps}~\cite{ovsjanikov2012functional} to tackle the above challenge. Originating from computer graphics, functional maps present a robust alternative to point-to-point correspondences~\cite{burghard2017embedding,kovnatsky2013coupled,nogneng2017informative}. They represent dense correspondences as linear mappings between function spaces, usually defined on 3D shapes. The key aspect of functional maps is their ability to capture deformations that align one manifold with another. Owing to their low-dimensional yet expressive nature, functional maps effectively incorporate global structures into the matching process. This approach provides a compelling solution to the challenges inherent in traditional point-wise correspondence methods.

Specifically, we improve zero-shot feature-based correspondence methods by transitioning from the pixel space to the function space, thereby enhancing the method's coherence and effectiveness. Traditional functional maps on manifolds rely on two geometric inputs: the Laplacian operator, which is crucial for computing the eigenfunction basis, and a local geometric descriptor, for the application of regularization losses. We adapt these components to the realm of images by employing visual features extracted from two distinct large vision models. Our approach diverges from traditional methods, which typically identify corresponding pixels between images through nearest neighbor search. Instead, we concentrate on optimizing a linear function map established on the eigenfunction basis defined by the first feature map, with the second feature map serving as a geometric regularizer. This process, notably unsupervised, marks a significant difference from conventional methods. Further augmenting our method's robustness, especially against occlusions, is the incorporation of a transformer module for tackling partial shape matching, as detailed in partial functional maps \etal~\cite{attaiki2021dpfm}. Such integration of functional map concepts with feature-based methods in image analysis represents a cohesive and logical advancement in tackling the challenges of correspondence tasks.

We evaluate our framework on dense correspondence across various base networks, demonstrating consistent enhancements in matching accuracy and other functional properties like smoothness compared to the traditional nearest neighbor search. We highlight the qualitative results of our approach on the challenging cases with significant shape variations, viewpoint changes, and occlusions. We further demonstrate our effectiveness on keypoint correspondences and object affordance map transfer, showcasing its versatility in diverse scenarios.

In summary, our primary contribution is a novel zero-shot framework designed to derive correspondence maps from pre-trained features. Central to our approach is the concept of optimizing a functional map that establishes a relationship between the entire image contents, moving away from the conventional method of direct pixel-to-pixel correspondence searches. Our experimental results, evaluated on various standard datasets, demonstrate that our method produces correspondences that are not only smoother and more accurate but also exhibit greater global coherence compared to previous efforts. We believe that our techniques effectively uncover the underlying correspondence capabilities of the large-scale backbone networks. We hope that our work will serve as an inspiration for future research in general-purpose object correspondence.

\section{Related Work}\label{sec:related_work}
\vspace{-.8em}

\paragraph{Emergent correspondence from vision models}
Deep image networks have demonstrated remarkable robustness to geometric transformations, such as rotation, scaling, and perspective changes, leading to the emergence of dense correspondences~\cite{dusmanu2019d2,ono2018lf,revaud2019r2d2,sarlin2020superglue,tyszkiewicz2020disk,yi2016lift}. These transformations, predominantly rigid in nature, have been a focal point in previous studies. The research by Amir \etal~\cite{amir2021deep} revealed that features extracted from DINOv1~\cite{caron2021emerging} not only act as effective dense visual descriptors but also naturally induce semantic correspondences without direct supervision. This capability is further amplified in its successor, DINOv2~\cite{oquab2023dinov2}. Beyond discriminative models, recent explorations have shown that generative models, such as diffusion models, also unveil emergent dense correspondences within their latent features~\cite{hedlin2023unsupervised,tang2023emergent,zhang2023tale}. Intriguingly, Zhang \etal~\cite{zhang2023tale} discovered that combining features from DINOv2~\cite{oquab2023dinov2} with those from Stable Diffusion~\cite{rombach2022high} significantly enhances correspondence quality. 

Our study highlights a crucial gap: existing methods lack structural awareness when computing correspondences by nearest-neighbor queries of per-pixel features. Here, we propose representing the correspondence map within a functional space, offering a novel approach to this challenge.

\paragraph{Semantic correspondence}
Semantic correspondence~\cite{liu2010sift} seeks to establish pixel-wise matches across objects differing in poses, appearances, deformations, or even categories. Traditional approaches generally involve three stages~\cite{truong2022probabilistic}: feature extraction, cost volume construction, and displacement field~\cite{truong2020glu,truong2020gocor,truong2021learning,truong2021warp} or parameterized transformation regression~\cite{jeon2018parn,kim2018recurrent,rocco2017convolutional,rocco2018end,seo2018attentive}. However, their reliance on smooth displacement fields or locally affine transformations hinders their ability to model complex object deformations or shape variations effectively.

Recent developments, inspired by the classical congealing method~\cite{learned2005data}, focus on aligning multiple objects within the same class using learning techniques like DINOv1 features~\cite{gupta2023asic,ofri2023neural} or GAN-synthesized data~\cite{peebles2022gan}. Despite their strong assumptions about data rigidity, these studies suggest that leveraging features and information from diverse tasks can enhance the quality of dense image correspondences. In our work, we further demonstrate that a structure-aware fusion of features learned from multiple tasks can significantly improve the quality of correspondence maps.

\paragraph{Functional maps}
Initially introduced by Ovsjanikov \etal~\cite{ovsjanikov2012functional} and further expanded by Aubry \etal~\cite{aubry2011wave}, functional maps offer a method to represent shape correspondences as linear transformations between spectral embeddings. This is achieved using compact matrices based on eigenfunction basis. Enhancements in accuracy, efficiency, and robustness have been realized in subsequent studies~\cite{kovnatsky2013coupled,huang2014functional,burghard2017embedding,nogneng2017informative}. Moving away from traditional methods dependent on hand-crafted features~\cite{sun2009concise,aubry2011wave}, recent developments have introduced various learning-based functional map frameworks. These utilize shape features learned via pairwise label supervision~\cite{litany2017deep}, geometric priors~\cite{halimi2019unsupervised,roufosse2019unsupervised}, or robust mesh features~\cite{sharp2022diffusionnet,cao2022unsupervised,li2022learning,donati2022deep}. While traditionally employed for full-shape correspondence, functional maps have also been adapted to handle partial correspondences~\cite{rodola2017partial,attaiki2021dpfm}, thus aligning more closely with real-world scenarios.

While functional maps are extensively explored for 3D shape representations like meshes and point clouds, their application to 2D images has been limited due to the ambiguous manifold structure of RGB-value representations~\cite{wang2013image,wang2014unsupervised}. Previous attempts at applying these maps to super-pixel image representations and utilizing their eigenfunctions as a basis~\cite{wang2013image,wang2014unsupervised} typically result in significant information loss. This is often due to the coarse nature of pre-segmentation in images and the resultant inconsistency in super-pixel representation. In our work, we address these challenges by using the entire image as input for a large vision model, ensuring a consistent initial representation and stable global structure during transformations by functional maps.

\section{Method}\label{sec:method}

\subsection{Preliminaries}

\paragraph{Functional map}
Originally introduced in Ovsjanikov \etal~\cite{ovsjanikov2012functional}, the functional map is a method for representing dense correspondences in the function space. This approach is based on the concept of mapping between function spaces defined on manifolds. Specifically, given two manifolds $\gM$ and $\gN$, we consider the spaces $\gF(\gM,\sR)$ and $\gF(\gN,\sR)$, each comprising all real-valued scalar functions on these manifolds, denoted as $\varphi^\gM:\gM\to\sR$ and $\varphi^\gN:\gN\to\sR$, respectively. 
We can express a bijective mapping $T: \gM \to \gN$ as a linear mapping between these function spaces, as follows:
\begin{equation}
    \label{eq1}
    \small
    T_F: \gF(\gM,\sR) \to \gF(\gN,\sR), \quad f \mapsto T_F(f).
\end{equation}

To compute these mappings effectively, we expand the function spaces $\gF(\gM,\sR)$ and $\gF(\gN,\sR)$ by introducing sets of basis functions, $\mathbf{\Phi}^\gM = \{\varphi_i^\gM\}$ and $\mathbf{\Phi}^\gN = \{\varphi_i^\gN\}$, for $\gM$ and $\gN$, respectively. Thus, any real-valued function $f \in \gF(\gM,\sR)$ can be represented as a linear combination of these basis functions: $f = \sum_i a_i \varphi_i^\gM$. Applying the operator $T_F$ to $f$ leads to the equation:
\begin{equation}
    \small
    T_F(f) = T_F\left(\sum_i a_i \varphi_i^\gM\right) = \sum_i a_i T_F(\varphi_i^\gM).
\end{equation}
Each transformed function $T_F(\varphi_i^\gM) \in \gF(\gN,\sR)$ can be further decomposed into a linear combination of $\varphi_j^\gN$. Hence, we have $T_F(\varphi_i^\gM) = \sum_j c_{ij}\varphi_j^\gN$, leading to:
\begin{equation}
    \small
    T_F(f) = \sum_i a_i \sum_j c_{ij} \varphi_j^\gN = \sum_h \sum_i a_i c_{ij} \varphi_j^\gN.
\end{equation}

\begin{figure*}[t]
    \centering
    \includegraphics[width=.9\linewidth]{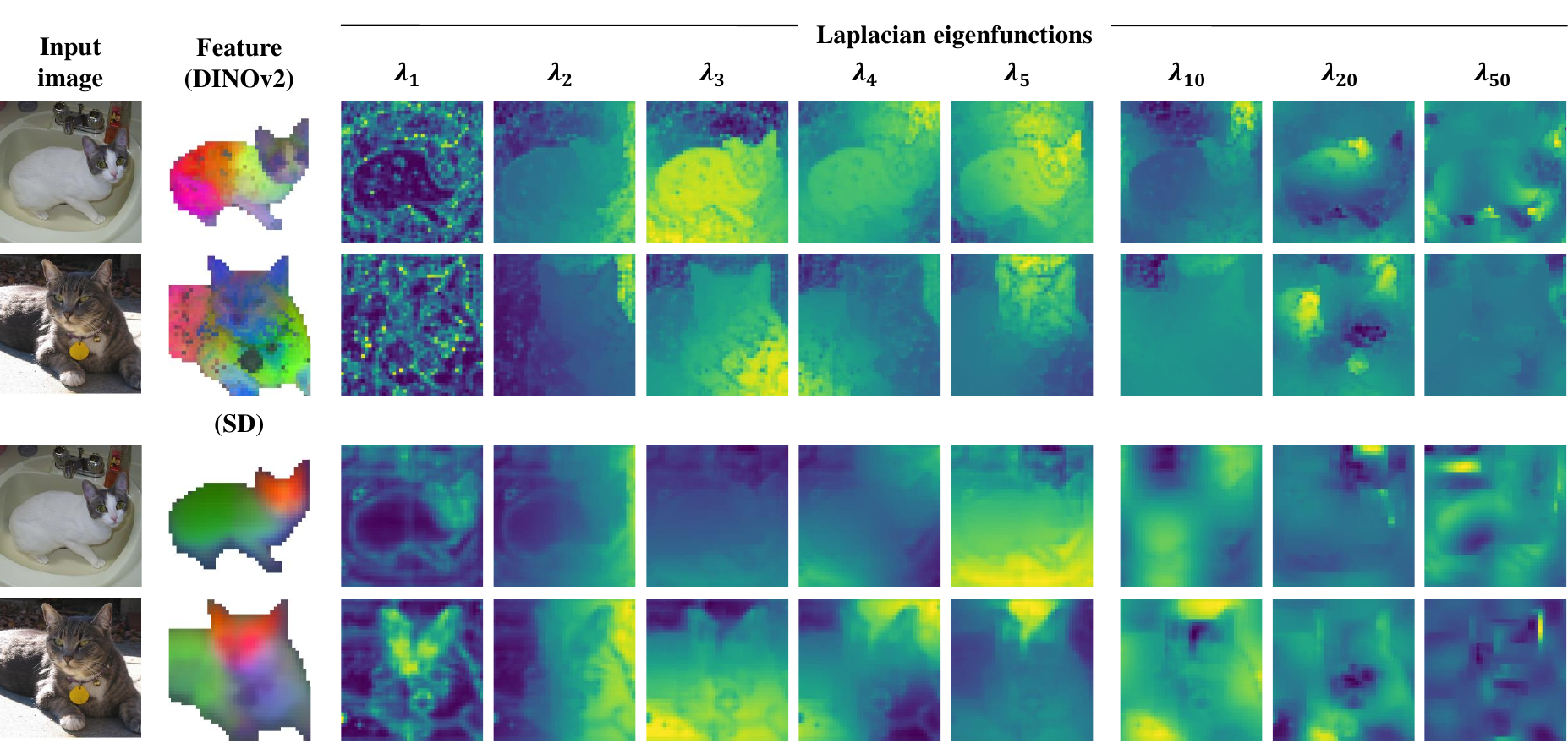}
    \vspace{-.8em}
    \caption{\textbf{Eigenfunctions of the image Laplacian.} We visualize the eigenfunctions of the graph Laplacian operator corresponding to the first 5 smallest eigenvalues $\lambda_1,\cdots,\lambda_5$ (low frequency) as well as $\lambda_{10}, \lambda_{20}, \lambda_{50}$ (high frequency).\vspace{-1em}}
    \label{fig:vis_eigenfunc}
\end{figure*}

For simplicity, the function $f$ is represented in a vector form with coefficients $\va = (a_1,a_2,\cdots)^t$. Consequently, the transformation $T_F$ on $\va$ is given by $T_F(\va) = \rmC\va$, where $\rmC$ is a matrix with elements $c_{ij}$, representing the $j$-th coefficient of $T_F(\varphi_i^\gM)$ in the basis $\{\varphi_j^\gN\}$.

To translate the functional map into point-to-point correspondences, we treat each point as a Dirac delta function in the function space. Specifically, this conversion from the functional to the point-wise map is executed via a nearest neighbor search between the rows of $\rmC\mathbf{\Phi}^\gM$ and $\mathbf{\Phi}^\gN$. A more detailed explanation of this process is available in the \SupplementaryMaterial.

\paragraph{Deep partial functional map}
The functional map framework, while adept at modeling perfect correspondence mappings between complete shapes~\cite{ovsjanikov2012functional}, faces challenges when applied to real-world data that often have missing data and noise. This has led to the development of partial functional maps, as discussed in~\cite{rodola2017partial,attaiki2021dpfm}.

The primary challenge in adapting functional maps to partial shapes is the disruption of core assumptions, such as manifold completeness and bijective mappings. Atta \etal~\cite{attaiki2021dpfm} address this challenge by introducing a feature refinement network, denoted as $g_\gR$, which enhances the robustness of partial functional maps against shape partiality.

Consider $M$ and $N$ as discretizations of the partial shapes $\gM$ and $\gN$, respectively. We construct a bipartite graph $(\mathcal{V}, \mathcal{E})$, with edges connecting every point $\rvx \in M$ to every point $\rvy \in N$. The refinement module inputs per-point features $F^M$ and $F^N$, and updates these features via message passing on the bipartite graph. This process employs an attention mechanism, formulated as
\begin{equation}
    \small
    m_{\epsilon \rightarrow i} = \sum_{j, (i,j) \in \mathcal{E}} \text{softmax}_j(q_i^{T}k_j/\sqrt{d}) v_j,
\end{equation}
and the final updated value of node $i$ is given by
\begin{equation}
    \small
    x_0 = x_0 + x_{\text{pos}}, \quad x_{i+1} = x_i + \text{MLP}([x_i \| m_{\epsilon \rightarrow i}]),
\end{equation}
where $x_{\text{pos}}$ represents the positional embedding, $[\cdot \| \cdot]$ denotes concatenation, and $\text{MLP}$ is a multilayer perceptron with ReLU activations and instance normalization. The refined features on the shape pair are denoted as $g_\gR(F^M)$ and $g_\gR(F^N)$.

To understand this message passing process, consider a region $\Omega$ exclusive to shape $M$ and absent in shape $N$. Let $F_\Omega$ denote a feature assignment function restricted to $\Omega$. When projecting these features onto the function basis, the functional map equation becomes:
\begin{equation}
    \small
    \rmC \varphi^M F_\Omega(M) = \varphi^N F_\Omega(N).
\end{equation}
This equation holds true if and only if $F_\Omega(\rvx) = 0$ implies $F_\Omega(\rvy) = 0$ for $\rvx \in M, \rvy \in N$. Hence, effective communication between the regions on $M$ and $N$ is crucial, enabling feature synchronization over overlapping regions while diminishing the influence of features outside these overlaps.

\subsection{Feature Consensus with Functional Maps}

An overview of our framework is depicted in \cref{fig:teaser}. Given a pair of images $M$ and $N$, our setup includes two distinct pixel-wise feature extraction networks, yielding two sets of features: $E^M, E^N$ and $F^M, F^N$. For instance, $E^M$ and $E^N$ might be DINOv2 features, while $F^M$ and $F^N$ could be Stable Diffusion features.

The primary objective is to derive a functional map $\rmC$ between the two function spaces $\gF(M,\sR)$ and $\gF(N,\sR)$. The core of our method involves using $E^M$ and $E^N$ to calculate the Laplacian eigenfunction basis and apply $F^M$ and $F^N$ for introducing regularizations in optimizing the functional map. In essence, our method optimizes the functional map derived from one set of features to achieve a ``consensus'' with the other set, providing a more comprehensive and robust mapping between the function spaces of the images.

\paragraph{Image Laplacian from visual features}
For an image feature of dimensions $(h, w)$, where $h$ is the height and $w$ is the width, we view it as a grid graph comprising $h \times w$ nodes; each node is connected to its four adjacent neighbors. However, a graph constructed naively would lack awareness of the image content, and its Laplacian eigenspaces would correspond to the conventional Fourier frequency space.

Instead, we assign weights to the graph edges based on the first set of image features $E^M$ and $E^N$. For two adjacent patches $\rvx$ and $\rvy$ in image $M$ (a similar definition applies for $N$), the weight of the edge between them is given by:
\begin{equation}
    \small
     \|e_{\rvx\rvy}\| = \exp\left(-\frac{\| E^M_\rvx - E^M_\rvy \|}{\sigma}\right),
\end{equation}
where $\sigma$ denotes the median of all the feature values. 

Next, we compute the graph Laplacian $\Delta_M$ and utilize its eigenfunctions as the basis. In alignment with previous research, we adopt a reduced function space spanned by the first 200 eigenfunctions. To compute the Laplacian eigen decompositions, we employ the LOBPCG algorithm, known for its efficiency. \cref{fig:vis_eigenfunc} presents examples of these Laplacian eigenfunctions.

\paragraph{Feature as function regularizer}
For the second set of features $F^M$ and $F^N$, we employ them as descriptor functions and impose a constraint on $\rmC$ such that $\rmC F^M \approx F^N$. 
Given the incompleteness of shape correspondences in image pairs, due for example to occlusion within the object and by other objects, we utilize the attention-based feature refinement network $g_\gR$ from deep partial functional maps~\cite{attaiki2021dpfm}. This network refines the features, which are then projected onto the function basis:
\begin{equation}
    \small
    \tilde{F}^M = \varphi^{M} g_\gR(F^M), \quad \tilde{F}^N = \varphi^{N} g_\gR(F^N).
\end{equation}
The descriptor-preserving loss applied to these refined features is formulated as:
\begin{equation}
    \small
    \gL_{\text{feat}} = \|\rmC \tilde{F}^M - \tilde{F}^N \|_2.
\end{equation}

To enhance the regularity of the functional map, our optimization objective incorporates two additional regularization terms. First, we integrate a compactness regularization into the functional map matrix:
\begin{equation}
    \small
    \gL_{\text{diag}} = \left( \left|\lambda^M_i - \lambda^N_j \right| c_{ij}\right)^2,
    \label{eq:loss_diag}
\end{equation}
where $\lambda^M_i$ and $\lambda^N_j$ represent the $i$-th and $j$-th eigenvalues of the graph Laplacian matrices $\Delta_M$ and $\Delta_N$, respectively. For images with similar spectral distributions of eigenvalues, minimizing $\gL_{\text{diag}}$ encourages a near-diagonal structure in $\rmC$. This regularization is based on the principle that eigenvalues' magnitudes are indicative of the frequencies of their corresponding eigenfunctions, and eigenfunctions with similar frequencies are more likely to correspond, as suggested by Huang \etal~\cite{huang2014functional}.

Next, we introduce a bijectivity constraint to the functional map:
\begin{equation}
    \small
    \rmC^{M\to N} \cdot \rmC^{N\to M} = \rmI.
\end{equation}
This can be interpreted as a special instance of the cycle-consistency regularization for image collections as in Wang \etal~\cite{wang2013image} when the number of images is two.

To implement this constraint, in line with Wang \etal~\cite{wang2013image}, we define two sets of optimizable latent bases: $\rmZ^M = \{Z^M_i\}$ and $\rmZ^N = \{Z^N_i\}$, corresponding to the function spaces $\gF(M,\sR)$ and $\gF(N,\sR)$ of both source and target images. The consistency loss is then defined as:
\begin{equation}
    \small
    \mathcal{L}_{\text{cons}} = \left\|\rmC \rmZ^M - \rmZ^N \right\|_2.
\end{equation}
To prevent degenerate solutions where $\rmZ^M$ and $\rmZ^N$ could be trivially zero, we introduce an additional constraint requiring both $\rmZ^M$ and $\rmZ^N$ to satisfy $\rmZ^t\rmZ = \rmI$.

Integrating all these components, our final optimization objective is:
\begin{equation}
    \small
    \begin{aligned}
        & \text{argmin}_{\rmC} \gL_{\text{feat}} + \lambda_{\text{diag}}\gL_{\text{diag}} + \lambda_{\text{cons}}\gL_{\text{cons}}, \\
        & \textit{s.t.}\quad{}(\rmZ^M)^t \rmZ^M = \rmI, (\rmZ^N)^t \rmZ^N = \rmI.
    \end{aligned}
    \label{eq:loss_all}
\end{equation}

\paragraph{Optimization}
We jointly optimize the weights of the image feature refinement network $g_\gR$, the functional map $\rmC$, and the latent basis $\rmZ^M$ and $\rmZ^N$ for the input image pair.
The full loss function is formulated as:
\begin{equation}
    \small
    \begin{aligned}
        \gL = &~ \gL_{\text{feat}} + \lambda_{\text{diag}}\gL_{\text{diag}} + \lambda_{\text{cons}}\gL_{\text{cons}} \\
              &~ + \lambda_Z \left( \text{tr}\left((\rmZ^M)^t \rmW \rmZ^M\right) + \text{tr}\left((\rmZ^N)^t \rmW \rmZ^N\right) \right)\\
              &~ + \lambda_{\text{reg}} \left(
                    \left\|(\rmZ^M)^t \rmZ^M - \rmI \right\|_2 +
                    \left\|(\rmZ^N)^t \rmZ^N - \rmI \right\|_2
                \right),
    \end{aligned}
\end{equation}
where $\rmW = \rmI + \rmC^t\rmC$. The terms $\text{tr}\left(\rmZ^t \rmW \rmZ\right)$ are variations of \cref{eq:loss_all} with $\rmZ^M$ and $\rmZ^N$ as the primary variables rather than $\rmC$, as discussed in Wang \etal~\cite{wang2013image}.

\section{Experiments}\label{sec:exp}
\vspace{-.5em}

\paragraph{Dataset}
We evaluate our method primarily on the TSS dataset~\cite{taniai2016joint}, comprising 400 image pairs from three subsets: FG3DCAR~\cite{lin2014jointly}, JODS~\cite{rubinstein2013unsupervised}, and PASCAL~\cite{hariharan2011semantic}, all of which include dense correspondence annotations. Additionally, we perform evaluations on the SPair-71k dataset~\cite{min2019spair}, which features sparse annotations of keypoint correspondences across 18 categories. For this dataset, we sample 20 pairs from each category for our analysis, following the prior work~\cite{zhang2023tale}.

\paragraph{Baselines}
Our comparison primarily focuses on emergent correspondences from various visual models and feature fusion techniques. We utilize feature extraction networks such as DINOv1 (ViT-S/8), DINOv2 (ViT-S/14 and ViT-B/14), and Stable Diffusion, which are prevalent and extensively researched in a wide range of visual perception tasks. In terms of feature fusion, we benchmark against the feature concatenation approach proposed by Zhang \etal~\cite{zhang2023tale}, testing different combinations of features. Additionally, we list other methods designed for image correspondence tasks that involve stronger supervision or task-specific designs.

\paragraph{Evaluation metrics}
For both dense and sparse correspondences, we adopt the Percentage of Correct Keypoints (PCK) metric~\cite{yang2012articulated} with a threshold of $\kappa\cdot\max(h,w)$, where $\kappa$ is a positive integer, and $(h,w)$ represents the image dimensions in the TSS dataset or the instance bounding-box dimensions in the SPair-71k dataset. Additionally, for dense correspondences on the TSS dataset, we assess spatial coherence using a smoothness metric~\cite{zhang2023tale}. This involves extracting a semantic flow (\ie, a 2D motion vector field from the source to the target image) and computing its first-order difference. In the case of sparse correspondences on the Spair-71k dataset, we further calculate the Mean Squared Error (MSE) on the keypoints to quantify mapping distortions.

\subsection{Dense Correspondence}\label{sec:exp:tss}

\input{tables/exp_tss}

Table~\ref{tab:exp_tss} presents the results of dense correspondences on the TSS dataset. Following \cite{zhang2023tale}, we majorly compare to other zero-shot unsupervised methods, among which we achieve the best performances. Specifically, we outperform Zhang \etal~\cite{zhang2023tale} with the same pair of features by utilizing the features in a more structure-aware manner. We also list as references the performances of fully supervised methods and unsupervised methods with task-specific training.

We also evaluate an ablated version of our framework by computing the basis functions and losses using the same set of features (the third and fourth rows from the last), which give significantly worse results compared to our full model. On the other side, it can still give better results than directly using one feature with nearest neighbor queries (for example, FMap DINOv2(basis) + DINOv2(loss) versus DINOv2-ViT-B/14). This shows that structure-awareness can naturally lead to better correspondences even without introducing any additional information.

\begin{figure*}[t]
    \centering
    \includegraphics[width=.9\linewidth]{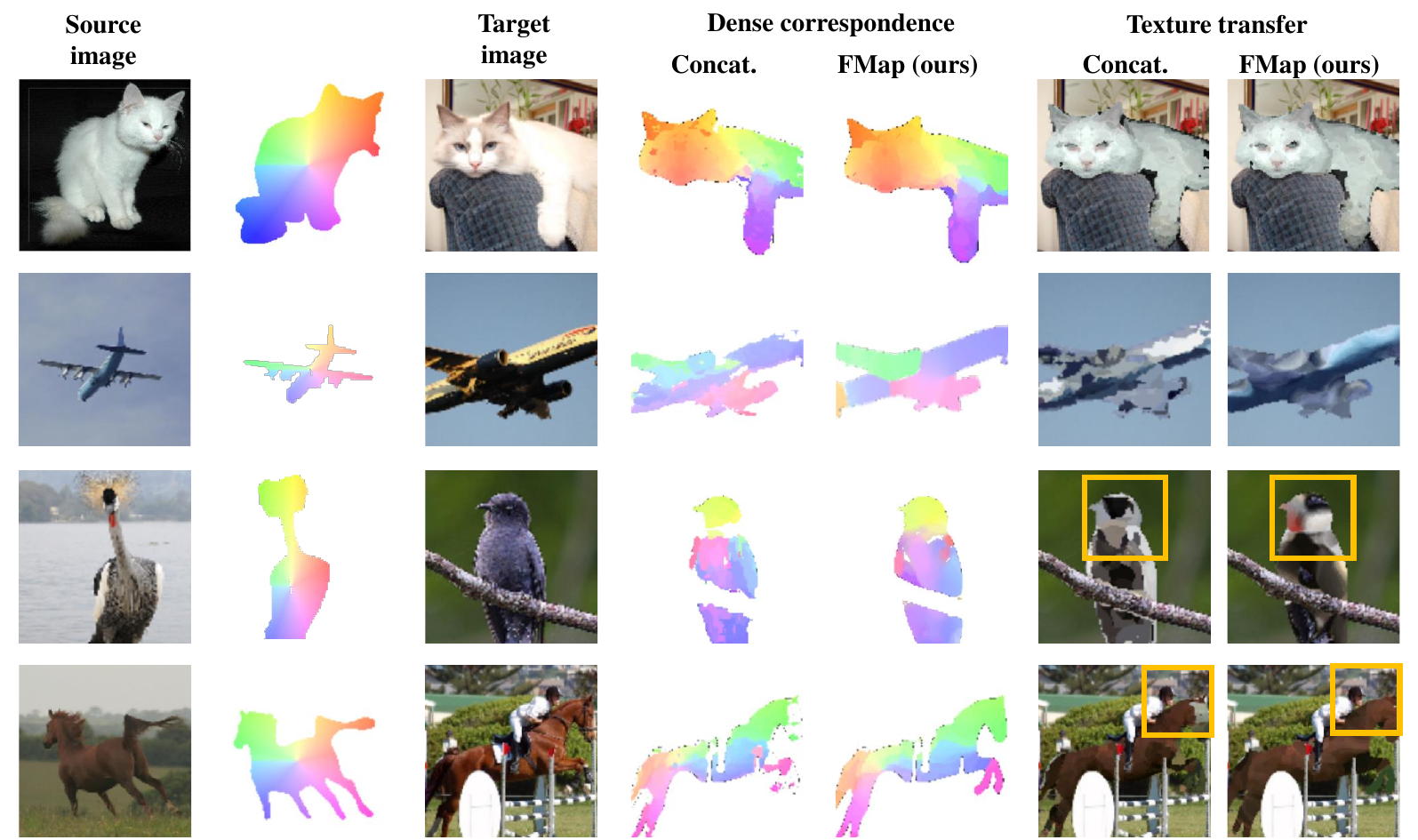}
    \vspace{-.8em}
    \caption{\textbf{Dense correspondences on SPair-71k~\cite{min2019spair} Image Pairs.} Each example displays pixel-wise mappings from source to target images in rainbow colors (second column for source coordinates, fourth and fifth columns for computed target coordinates) and color transfers (last two columns). Specifically, we demonstrate the challenging examples including significant viewpoint changes (first and second row), shape variations (first and third row), and occlusions (third row). Our framework achieves more consistent mappings with its global structure-awareness.\vspace{-1.5em}}
    \label{fig:exp_dense_corr}
\end{figure*}

\cref{fig:exp_dense_corr} shows the qualitative results of dense correspondences computed with the DINOv2-ViT-B/14 and Stable Diffusion networks.
We compare side-by-side the feature fusion results using pre-normalized concatenation~\cite{zhang2023tale} and our method.
In all these examples, our framework provides smoother and more consistent mappings with its global structure-awareness. 
Specifically, we highlight two challenging examples: the airplanes in the second row with large camera-view changes, and the birds in the third row with large shape variations as well as occlusions.
We also visualize the matrices for the linear functional maps in \cref{fig:vis_fmap_matrix}.

\input{tables/exp_different_backbones}

\paragraph{Feature fusion with different networks}
\cref{tab:exp:different_backbones} presents the accuracy and smoothness of correspondences derived from features of various network backbones. When compared to using individual features or their concatenation~\cite{zhang2023tale}, our functional-map-based framework demonstrates superior results in both metrics across all tested configurations.

\input{tables/exp_different_layers}

\paragraph{Feature fusion with different layers}\label{sec:exp:diflayer}
\cref{tab:exp:different_layers} presents the results of fusing features from different layers within the same network. Our experiments involve layers 9 and 11 of DINOv2-ViT-S/14 and DINOv2-ViT-B/14. In all tested setups, our framework demonstrates superior performance compared to baseline methods.

\input{tables/exp_layer_choice}

Additionally, a comparative analysis was performed on the choice of layers for DINOv2-ViT-B/14, specifically by fusing the features of layer 11 with those of layers 8, 9, 10, and layer 11 tokens. The results, as depicted in \cref{tab:exp:layer_choice}, indicate that our functional map approach consistently surpasses both raw and concatenated features across all layer combinations.
We also observed that greater feature distinction occurs when the two layers are more distant from each other. Our framework effectively leverages this distinction, extracting better correspondences by integrating their information. As shown in \cref{tab:exp:layer_choice}, optimal performance in EPE is achieved using features from layers 8 and 11.

\subsection{More Results}\label{sec:exp:spair}

\begin{figure*}[t]
\vspace{-1em}
    \centering
    \begin{subfigure}[t]{\linewidth}
        \centering
        \includegraphics[width=.9\linewidth, trim=0.5cm 2.5cm 0cm 2.5cm, clip]{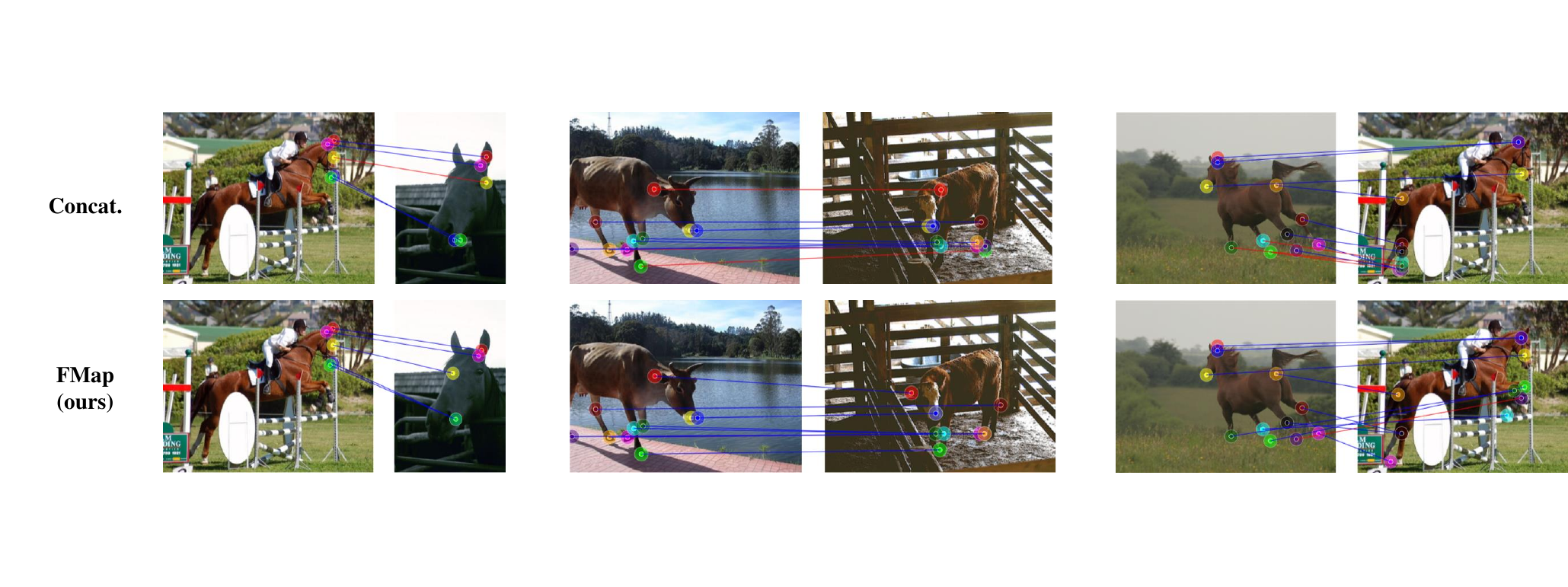}
        \caption{\scriptsize Image pairs with similar geometric properties. (a) The baseline method incorrectly maps (a) the right ear of the horse to the left ear, (b) the right ear of the cow to the left ear, and (c) a point corresponding to the front feet of the horse to the hind feet.}
        \label{fig:exp_sparse_corr_1}
    \end{subfigure}
    \begin{subfigure}[t]{\linewidth}
        \centering
        \vspace{.5em}
        \includegraphics[width=.9\linewidth, trim=0.5cm 2.5cm 0cm 2.5cm, clip]{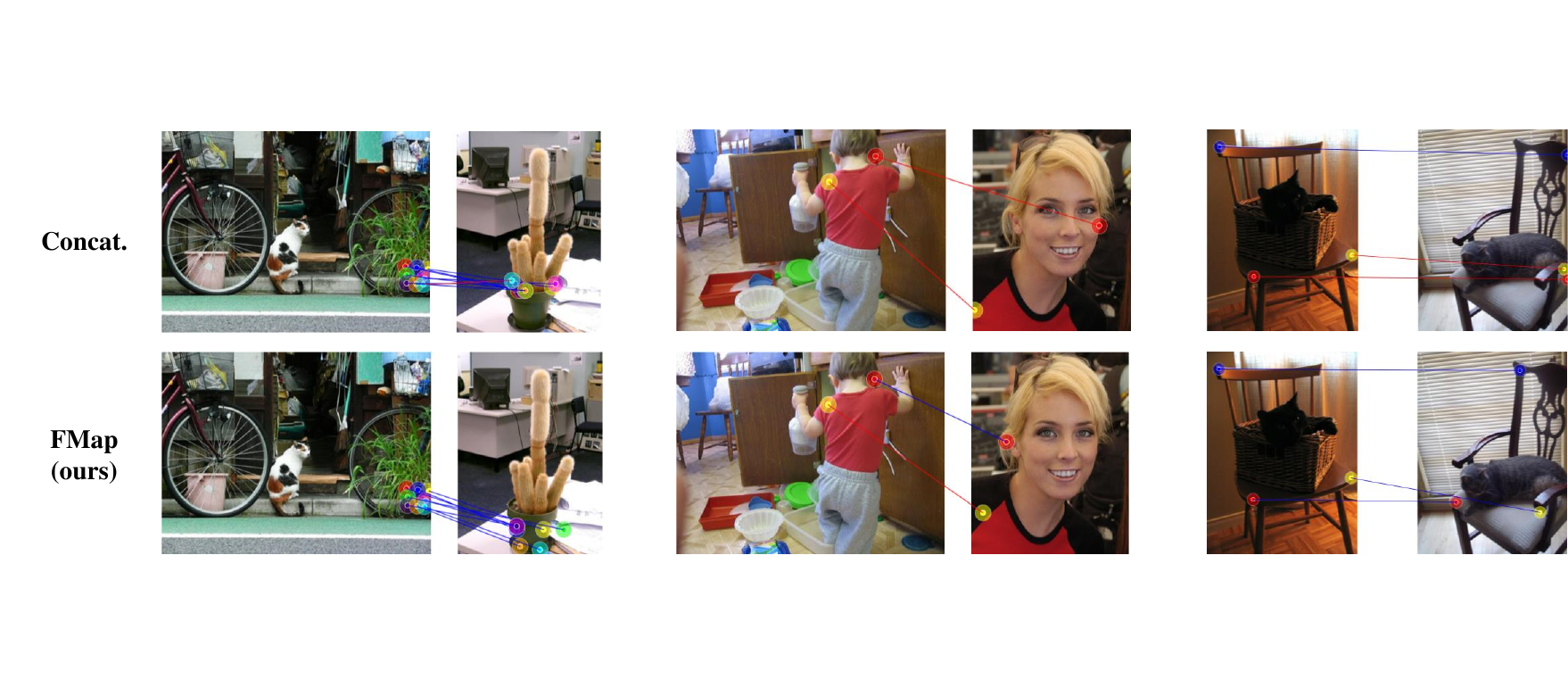}
        \caption{\scriptsize Image pairs with significant differences in shapes and viewpoints. The baseline method incorrectly maps (a) all points on the pot to the plant, (b) a point on the child's ear to the woman's cheek, and (c) a point at the seat corner to another chair's armrest.}
        \label{fig:exp_sparse_corr_2}
    \end{subfigure}
    \vspace{-.5em}
    \caption{\textbf{Sparse keypoint correspondences on SPair-71k~\cite{min2019spair} image pairs.} Correct matches are connected with {\color{blue}blue lines} and incorrect matches with {\color{red}red lines}.\vspace{-1.5em}}
    \label{fig:exp_sparse_corr}
\end{figure*}

\input{tables/exp_spair}

\paragraph{Keypoint correspondence}
\cref{tab:exp:spair} presents the results for sparse keypoint correspondences on SPair-71k~\cite{min2019spair}. Compared to feature concatenation \cite{zhang2023tale}, our method demonstrates comparable or higher PCK (with different thresholds) and exhibits lower MSE errors. Note that the selected keypoints are extremely sparse on the images, which could potentially introduce sampling biases compared to evaluations of dense correspondences.

\cref{fig:exp_sparse_corr} showcases qualitative keypoint matching results. Our method is compared side-by-side with results obtained using feature concatenation, where our approach consistently demonstrates robustness in these challenging scenarios and effectively captures the geometric properties of the features. \cref{fig:exp_sparse_corr_1} further illustrates the effectiveness of our method in scenarios where the target image contains many similar points, like the legs of a horse. In contrast, the baseline struggles to capture the global structure, often leading to mappings of similar but incorrect points.

\begin{figure*}[t!]
    \centering
    \vspace{-.8em}
    \includegraphics[width=.9\linewidth, trim=0.95cm 3cm 0.95cm 3cm, clip]{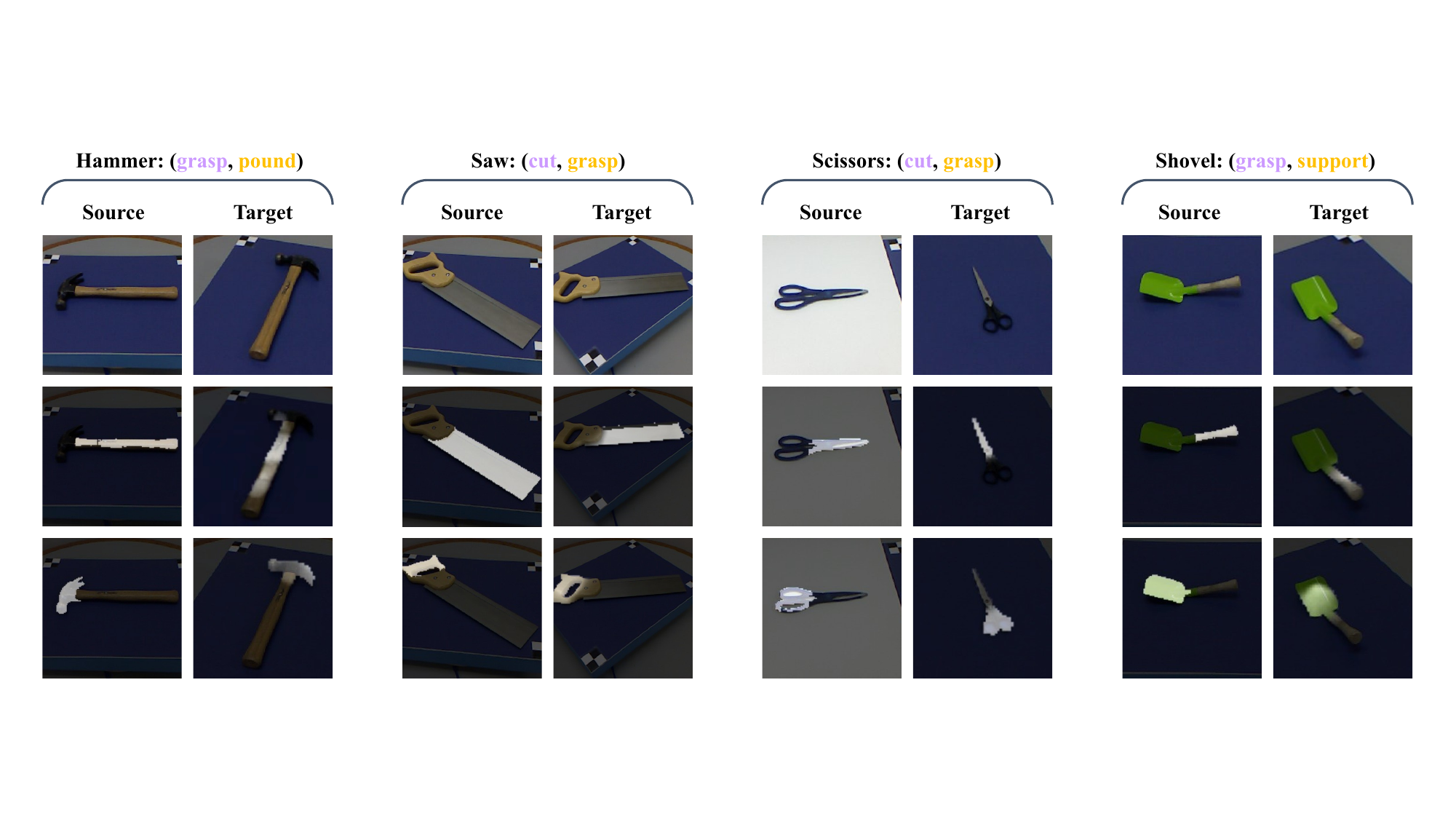}
    \vspace{-.8em}
    \caption{\textbf{Transferring tool affordances represented as heat maps.} We treat affordance heat maps as functions defined on the source and the target image. By optimizing the functional map between the source and the target, we manage to transfer the function after applying the functional map to it directly following \cref{eq1}. We employ features from DINOv2-ViT-B/14 and Stable Diffusion to compute the functional maps in this experiment.\vspace{-1em}}
    \label{fig:app_affordance}
\end{figure*}

\paragraph{Affordance transfer}
We further showcase an application of our method in transferring tool affordances between images from the RGB-D Part Affordance Dataset~\cite{myers2015affordance}. This dataset features different types of affordances annotated on each object, represented as heat maps. \cref{fig:app_affordance} illustrates our results in transferring these affordance heat maps. Such distributional functions across pixels pose a challenge to raw pixel-wise maps due to the potential distortion of their overall structure during interpolation. However, these functions can be naturally modeled with functional maps, as our approach demonstrates.

\begin{figure*}[t]
    \centering
    \includegraphics[width=.9\linewidth]{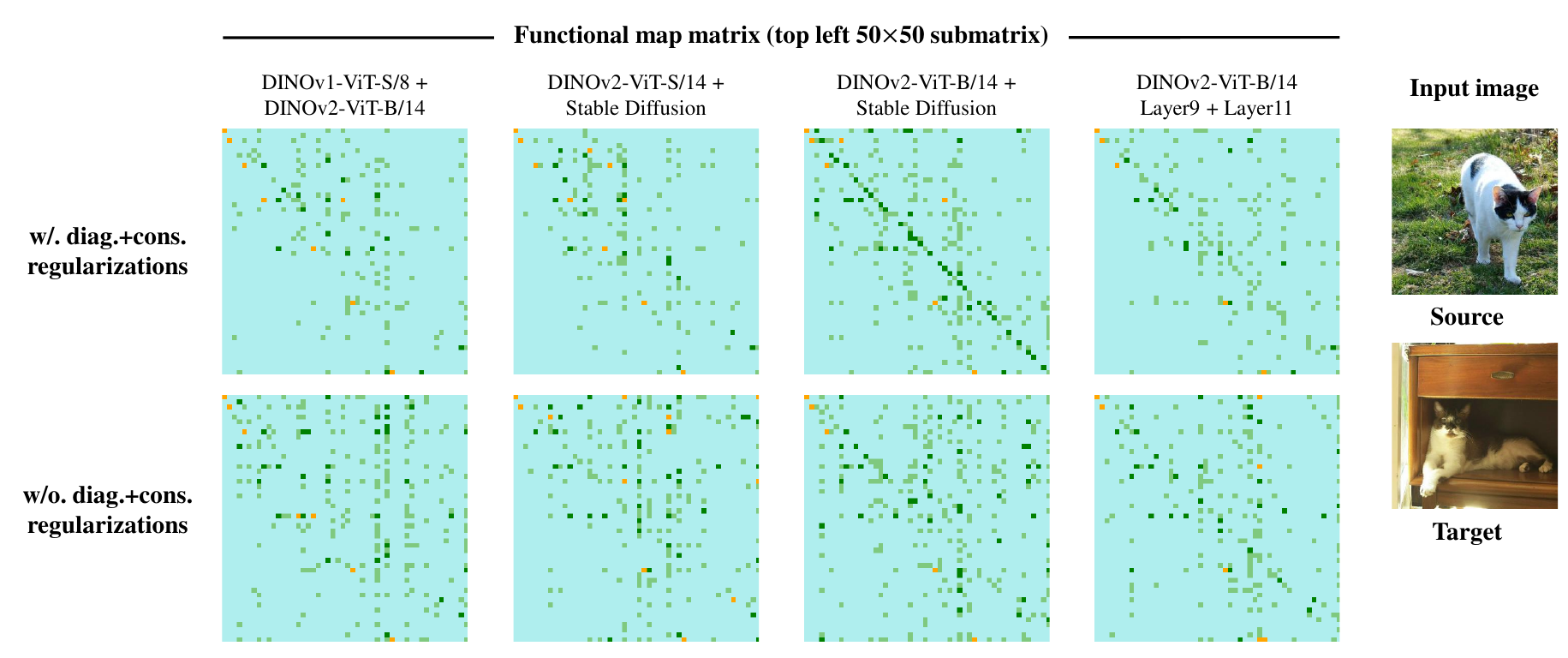}
    \vspace{-1em}
    \caption{\textbf{Functional map matrices with and without regularization losses.} Enforcing the compactness loss (\cref{eq:loss_diag}) centers the non-zero matrix entries around the diagonals to match the function basis of similar frequencies.\vspace{-.5em}}
    \label{fig:vis_fmap_matrix}
\end{figure*}

\input{tables/exp_ablation}

\paragraph{Ablation Studies}\label{sec:exp:ablation}
In addition to the feature ablations shown in \cref{tab:exp_tss} and discussed in \cref{sec:exp:tss}, we also present an ablation on the regularization terms for the functional map optimization. \cref{tab:exp_ablation} shows the results optimized with different regularization losses. The diagonality and consistency regularizations greatly improve the accuracy of the mapping.
\cref{fig:vis_fmap_matrix} visualizes the functional map matrics with and without the regularizations. The near-diagonal mappings are preferred because they match the function basis with similar frequencies.

\section{Discussions}
\vspace{-.5em}

As shown in \cref{sec:exp:diflayer}, our functional map framework effectively integrates features from different network layers. This integration, particularly from just two distinct layers, outperforms the conventional approach of using same-layer features or naively concatenating different features. This finding opens up promising avenues for enhancing the generalization capabilities of large-scale vision models \emph{without additional fine-tuning}.
Our method also aligns with the high-level principles of the Adaptive Mixture of Local Experts (MoE)~\cite{jacobs1991adaptive}. It can be seen as each ``expert'' in our framework reaching a consensus on the geometric properties of features, akin to a specialized gating network within the MoE paradigm.

Moreover, the interpretability of learned features in the functional map framework is crucial, particularly in domains like medical imaging or autonomous systems. Our approach, as shown in \cref{fig:exp_dense_corr}, enables impressive image editing outcomes without generative models. This leads to the intriguing possibility of combining our method with generative models to enhance image quality.

In summary, the interpretability, performance, and potential for integration with generative models position our functional map framework as a versatile tool in computer vision. Its ongoing exploration and application in real-world scenarios further underscore its importance in the field of visual understanding and synthesis.

\vspace{-.5em}
\section{Conclusions}
\vspace{-.8em}

The emergence of correspondences from large-scale vision models not explicitly trained for this task is noteworthy. While nearest-neighbor analyses provide a direct exploration, they overlook the structure inherent not only in the image contents but also in the model features. Our work leverages this embedded structure via functional maps, aiming to generate point-wise accurate and globally coherent correspondences. Despite its simplicity, it significantly enhances the matching results, both qualitatively and quantitatively, with zero-shot inference on image pairs without additional supervision or task-specific training. While the core concepts of our approach are rooted in 3D shape correspondence literature from graphics~\cite{ovsjanikov2012functional}, our implementation using deep feature-based functional maps bridges this area with cutting-edge vision research. We believe this connection will inspire broader possibilities in image correspondence problems.

\paragraph{Limitations and future work}
The structure-awareness of functional maps relies on the manifold assumption of its underlying domain, making our current framework more suitable for object-centric images than complex scenes with diverse compositionalities. Examples of the latter include matching a horse to a herd of horses or matching two indoor scenes. However, this issue might be addressed using additional image segmentation techniques that decompose the image into objects and parts, or by exploring matches between quotient spaces.

%
%
\bibliographystyle{splncs04}
\bibliography{reference_header,reference}

\appendix
\renewcommand\thefigure{A\arabic{figure}}
\setcounter{figure}{0}
\renewcommand\thetable{A\arabic{table}}
\setcounter{table}{0}
\renewcommand\theequation{A\arabic{equation}}
\setcounter{equation}{0}
\pagenumbering{arabic}
\renewcommand*{\thepage}{A\arabic{page}}
\setcounter{footnote}{0}

\input{appendix}

\end{document}

%% file: configs.tex
\definecolor{darkred}{rgb}{0.6, 0.1, 0.05}

\newcommand{\SupplementaryMaterial}{{\color{darkred} supplementary material}}

\definecolor{TabPurple}{HTML}{EECFD4}
\definecolor{TabOrange}{HTML}{FFF275}

\newcolumntype{x}{>{\columncolor{LightCyan1}}c}
\newcolumntype{y}{>{\columncolor{MistyRose}}c}

%% file: math_commands.tex
\usepackage{amsmath,amsfonts,bm}

\def\eqref#1{equation~\ref{#1}}

\def\1{\bm{1}}

\def\rvx{{\mathbf{x}}}
\def\rvy{{\mathbf{y}}}

\def\rmC{{\mathbf{C}}}

\def\rmI{{\mathbf{I}}}

\def\rmW{{\mathbf{W}}}

\def\rmZ{{\mathbf{Z}}}

\def\va{{\bm{a}}}

\DeclareMathAlphabet{\mathsfit}{\encodingdefault}{\sfdefault}{m}{sl}
\SetMathAlphabet{\mathsfit}{bold}{\encodingdefault}{\sfdefault}{bx}{n}

\def\gF{{\mathcal{F}}}

\def\gL{{\mathcal{L}}}
\def\gM{{\mathcal{M}}}
\def\gN{{\mathcal{N}}}

\def\gR{{\mathcal{R}}}

\def\sR{{\mathbb{R}}}



%% file: tables/exp_tss.tex
\begin{table}[t]
    \centering
    \footnotesize
    \setlength{\tabcolsep}{3pt}
    \caption{\textbf{Results for dense correspondences on TSS \cite{taniai2016joint}.} The baselines are classified into three categories based on their training setups: supervised, unsupervised with task-specific designs, and zero-shot methods without task- or dataset-specific designs. * indicates backbones fine-tuned on this dataset.\vspace{-.8em}}
    \label{tab:exp_tss}
    \resizebox{\textwidth}{!}{%
    \begin{tabular}{llxxxy}
    \toprule
        \textbf{Setting} & \textbf{Method} & \textbf{FG3DCar} & \textbf{JODS} & \textbf{Pascal} & \textbf{Avg.} \\
    \midrule
        \multirow{3}{*}{Supervised}
        & SCOT \cite{liu2020semantic} & 95.3 & 81.3 & 57.7 & 78.1 \\
        & CATs$^*$ \cite{cho2021cats} & 92.1 & 78.9 & 64.2 & 78.4 \\
        & PWarpC-CATs$^*$ \cite{truong2022probabilistic} & 95.5 & 85.0 & 85.5 & 88.7 \\
    \midrule
        \multirow{4}{*}{\makecell[l]{Unsupervised\\task-specific}}
        & CNNGeo \cite{rocco2017convolutional} & 90.1 & 76.4 & 56.3 & 74.4 \\
        & PARN \cite{jeon2018parn} & 89.5 & 75.9 & 71.2 & 78.8 \\
        & GLU-Net \cite{truong2020glu} & 93.2 & 73.3 & 71.1 & 79.2 \\
        & Semantic-GLU-Net \cite{truong2021warp} & 95.3 & 82.2 & 78.2 & 85.2 \\
    \midrule
        \multirow{8}{*}{\makecell[l]{Unsupervised\\zero-shot}}
        & DINOv1-ViT-S/8 \cite{amir2021deep} & 68.7 & 44.7 & 36.7 & 52.7 \\
        & DINOv2-ViT-B & 81.2 & 68.4 & 51.5 & 69.4 \\
        & Stable Diffusion (SD) & 92.1 & 62.6 & 48.4 & 72.5 \\
        & Concat. DINOv2 + SD \cite{zhang2023tale} &  92.9&73.8  &59.6  & 78.7 \\
        & FMap DINOv2(basis) + DINOv2(loss) & 83.5 & 69.2 & 52.7 & 71.0 \\
        & FMap SD(basis) + SD(loss) &  80.0& 63.4& 51.5 & 67.8 \\
        & FMap DINOv2(basis) + SD(loss) \textbf{(ours)} & 84.8 & 70.4 & 53.5 & 72.2\\
        & FMap DINOv2(loss) + SD(basis) \textbf{(ours)} & \textbf{93.1} & \textbf{74.0} & \textbf{59.9} & \textbf{78.9}\\
    \bottomrule
    \end{tabular}
    }
    \vspace{-1em}
\end{table}

%% file: tables/exp_different_backbones.tex
\begin{table}[t]
    \centering
    \footnotesize
    \caption{\textbf{Fusing the features from different networks.}\vspace{-.8em}}
    \label{tab:exp:different_backbones}
    \setlength{\tabcolsep}{3pt}
    \begin{tabular}{llxxxx}
        \toprule
        \multicolumn{2}{c}{\textbf{Method}}
        & PCK0.05$\uparrow$ & PCK0.1$\uparrow$ & EPE$\downarrow$ & Smth.$\downarrow$ \\
        \midrule
        DINOv1-ViT-S/8 & raw
        & 53.9& 76.8 & 46.1 & 12.90 \\
        DINOv2-ViT-S/14 & raw
        & 69.6 & 85.0 & 30.8 & ~~7.98 \\
        DINOv2-ViT-B/14 & raw
        & 69.4 & 87.8 & 30.9& 10.46 \\
        Stable Diffusion (SD) & raw
        & 72.5& 83.8 & 37.5& ~~6.41 \\
        \midrule
        \multirow{2}{*}{\makecell[l]{DINOv1-ViT-S/8\\+ DINOv2-ViT-B/14}}
        & Concat. \cite{zhang2023tale}
        & 69.9 & 88.1 & 31.0 & 10.33 \\
        & FMap \textbf{(ours)}
        & \textbf{72.2} & \textbf{90.3} & \textbf{27.7} & ~~\textbf{7.95} \\
        \midrule
        \multirow{2}{*}{DINOv2-ViT-S/14 + SD}
        & Concat. \cite{zhang2023tale}
        & \textbf{78.1} & 89.9 & 27.5 & ~~6.58 \\
        & FMap \textbf{(ours)}
        & 71.5 & \textbf{90.0} & \textbf{26.3} & ~~\textbf{6.47} \\
        \midrule
        \multirow{2}{*}{DINOv2-ViT-B/14 + SD}
        & Concat. \cite{zhang2023tale}
        & 78.7 & 90.7 & 26.4& ~~6.81 \\
        & FMap \textbf{(ours)}
        & \textbf{78.9} & \textbf{91.1} & \textbf{26.1} & ~~\textbf{5.74} \\
        \bottomrule
    \end{tabular}%
\end{table}

%% file: tables/exp_different_layers.tex
\begin{table}[t]
    \centering
    \footnotesize
    \caption{\textbf{Fusing the features from different layers of the same network.}\vspace{-.8em}}
    \label{tab:exp:different_layers}
    \setlength{\tabcolsep}{3pt}
    \begin{tabular}{llxxxx}
        \toprule
        \textbf{Backbone} & \textbf{Method} & PCK0.05$\uparrow$ & PCK0.1$\uparrow$ & EPE$\downarrow$ & Smth.$\downarrow$ \\
        \midrule
        \multirow{4}{*}{DINOv2-ViT-S/14}
        & Layer9
        & 67.2 & 84.8 & 36.5 & ~~9.64 \\
        & Layer11
        & 70.8 & 88.1 & 31.0 & ~~9.25 \\
        & Concat. \cite{zhang2023tale}
        & 70.5 & 88.1 &31.0 & ~~9.25 \\
        & FMap \textbf{(ours)}
        & \textbf{70.8} & \textbf{89.1} & \textbf{29.1} & ~~\textbf{6.60} \\
        \midrule
        \multirow{4}{*}{DINOv2-ViT-B/14}
        & Layer9
        & 57.2 & 85.4 & 34.5 & 10.66  \\
        & Layer11
        & 69.4 & 87.8 & 30.9 & 10.46 \\
        & Concat. \cite{zhang2023tale}
        & 70.0 & 87.9 & 30.9 & 10.24 \\
        & FMap \textbf{(ours)}
        & \textbf{70.6} & \textbf{89.8} & \textbf{25.9} & ~~\textbf{8.27} \\
        \bottomrule
    \end{tabular}%
    \vspace{-1em}
\end{table}

%% file: tables/exp_layer_choice.tex
\begin{table}[t]
    \centering
    \footnotesize
    \caption{\textbf{Results on different layer choices for feature fusion.} This experiment involves DINOv2-ViT-B/14, wherein its layer 11 features (values) are fused with layers 8, 9, 10, and layer 11 tokens, respectively.\vspace{-.8em}}
    \setlength{\tabcolsep}{3pt}
    \begin{tabular}{lxxyyxxyy}
        \toprule
        \multirow{2}{*}{\textbf{Method}} & \multicolumn{2}{x}{Layer 8} & \multicolumn{2}{y}{Layer 9} & \multicolumn{2}{x}{Layer 10} & \multicolumn{2}{y}{Layer 11 token} \\
        & EPE$\downarrow$ & Smth.$\downarrow$ & EPE$\downarrow$ & Smth.$\downarrow$ & EPE$\downarrow$ & Smth.$\downarrow$ & EPE$\downarrow$ & Smth.$\downarrow$ \\
        \midrule
        Raw \cite{amir2021deep} &
        59.1& 16.10 & 56.8 &16.06  & 56.8 &15.40  & 53.3 &13.20  \\
        Concat. \cite{zhang2023tale} &
        53.5&  14.80&  55.4& 13.90 & 56.7 &16.70  & 55.3 & 16.10 \\
        FMap \textbf{(ours)} &
        \textbf{41.8} & \textbf{11.95} & \textbf{45.2} & ~~\textbf{9.52} & \textbf{41.9} & \textbf{12.43} & \textbf{45.3} & \textbf{10.65} \\
        \bottomrule
    \end{tabular}
    \label{tab:exp:layer_choice}
\end{table}

%% file: tables/exp_spair.tex
\begin{table}[t]
    \centering
    \footnotesize
    \setlength{\tabcolsep}{3pt}
    \caption{\textbf{Results for sparse keypoint correspondences on SPair-7k~\cite{min2019spair}.} All results in this experiment are with the DINOv2-ViT-B/14 backbone.\vspace{-.8em}}
    \label{tab:exp:spair}
    \begin{tabular}{lxxx}
        \toprule
        \textbf{Method} & PCK@0.1$\uparrow$& PCK@0.2$\uparrow$ & MSE$\downarrow$ \\
        \midrule
        DINOv2 &52.3 & 68.0 & 105.0 \\
        Stable Diffusion & 51.2& 64.1 & 120.5 \\
        Concat. \cite{zhang2023tale} & \textbf{57.2}&72.2 & ~~97.2 \\
        FMap \textbf{(ours)} & 55.3&\textbf{72.6} & ~~\textbf{88.0} \\
        \bottomrule
    \end{tabular}
\end{table}

%% file: tables/exp_ablation.tex
\begin{table}[t]
    \centering
    \footnotesize
    \setlength{\tabcolsep}{3pt}
    \caption{\textbf{Ablation on the loss terms.} All results in the experiment are with DINOv2-ViT-B/14 and Stable Diffusion on the SPair-71k dataset.\vspace{-.8em}}
    \label{tab:exp_ablation}
    \begin{tabular}{lxxx}
        \toprule
        \textbf{Loss} & PCK@0.1$\uparrow$ &PCK@0.2$\uparrow$ & MSE$\downarrow$ \\
        \midrule
        $\gL_{\text{feat}}$ (no regularization) & 44.6& 65.5 & ~~95.3\\
        $\gL_{\text{feat}}+\gL_{\text{diag}}$&52.9 & 69.5 & ~~97.9 \\
        $\gL_{\text{feat}}+\gL_{\text{cons}}$ &52.8& 69.7 & 100.3\\
        $\gL_{\text{feat}}+\gL_{\text{diag}}+\gL_{\text{cons}}$ (full loss) &\textbf{55.3}& \textbf{72.6} & ~~\textbf{88.0} \\
        \bottomrule
    \end{tabular}
    \vspace{-1em}
\end{table}

%% file: appendix.tex
\section{Implementation Details}

We implemented the image feature functional map in approximately 400 lines of Python code, utilizing PyTorch v1.9.1 and CUDA 11.1.

\paragraph{Graph Laplacian construction from images}
Our method employs PyG (PyTorch Geometric), a comprehensive library designed for deep learning on graphs. We utilize its capabilities to refine edge weights based on feature differences and to compute the graph's Laplacian matrix.

\paragraph{Computation and resources}
All experiments were carried out using the PyTorch framework on a 64-bit machine, equipped with a single NVIDIA GeForce RTX 3090 GPU.

\paragraph{Image preprocessing}
For the SPair-71k dataset, we apply image cropping using the provided bounding boxes, followed by resizing both source and target images to a uniform scale. For the TSS dataset, which primarily features object-centric images, such cropping is generally not necessary.

\section{Functional Map Details}\label{sec:supp:functionalmap}

\paragraph{Functional map to point-wise map}
To translate the functional representation back to the original mapping, we determine a corresponding point \(y \in \mathcal{N}\) for each \(x \in \mathcal{M}\). In the Laplacian basis, the matrix $\mathbf{\Phi}^\gM$ represents the Laplacian eigenfunctions of \(\mathcal{M}\), with columns and rows corresponding to points and eigenfunctions, respectively. The image of all delta functions at \(\mathcal{M}\)'s points is represented by $\rmC\mathbf{\Phi}^\gM$.

\paragraph{Function distance in spectral domain}
For functions \(g_1\) and \(g_2\) on \(N\) with spectral coefficients \(b_1\) and \(b_2\), the equation \(\sum_i (b_{1i} - b_{2i})^2 = \int_N (g_1(y) - g_2(y))^2 \mu(y)\) holds. This equates the coefficient vector distances to the \(L_2\) difference between the functions.

\paragraph{Correspondence estimation}
Establishing point correspondences effectively involves finding the nearest neighbor in $\mathbf{\Phi}^\gN$ for each point in $\rmC\mathbf{\Phi}^\gM$.

\paragraph{Descriptor preservation}
For point descriptors represented by functions \(f\) and \(g\), our mapping aims to retain descriptor qualities. In multidimensional descriptor scenarios, \(f(x) \in \mathbb{R}^k\) for each \(x\), we establish \(k\) constraints for scalar functions, addressing each descriptor dimension individually.

\paragraph{Hyperparameters}
Hyperparameters were carefully chosen, with the consistency loss parameter ($\lambda_{\text{cons}}$) set around 1e-3. The regularization loss parameters $\lambda_{\text{diag}}$, $\lambda_{\text{reg}}$, and $\lambda_{Z}$ were set to 5 and 1, respectively, balancing regularization strength and model expressiveness.

\section{Per-Category Results}\label{sec:exp_percat}

\begin{table}[ht!]
    \centering
    \footnotesize
    \caption{\textbf{Per-category results on TSS (\cref{tab:exp:different_backbones,tab:exp:different_layers}, \cref{sec:exp:tss}).
    }}
    \vspace{-.5em}
    \setlength{\tabcolsep}{2pt}
    \resizebox{\linewidth}{!}{%
        \begin{tabular}{lxxxxyyyyxxxxyyyy}
            \toprule
            \multirow{2}{*}{\textbf{Method}}
            & \multicolumn{4}{x}{PCK@0.05$\uparrow$}
            & \multicolumn{4}{y}{PCK@0.10$\uparrow$}
            & \multicolumn{4}{x}{EPE$\downarrow$}
            & \multicolumn{4}{y}{Smth.$\downarrow$} \\
            & \textbf{FG3DCar} & \textbf{JODS} & \textbf{Pascal}  & \textbf{Avg.}
            & \textbf{FG3DCar} & \textbf{JODS} & \textbf{Pascal}  & \textbf{Avg.}
            & \textbf{FG3DCar} & \textbf{JODS} & \textbf{Pascal}  & \textbf{Avg.}
            & \textbf{FG3DCar} & \textbf{JODS} & \textbf{Pascal}  & \textbf{Avg.}\\
            \midrule
            DINOv1-ViT-S/8 &68.7&44.7&36.7&53.9&89.1&69.1 &62.5 &76.8&40.8&35.8 &61.4&46.1&8.02&18.14&17.40&12.90
            \\
            DINOv2-ViT-S/14
            &83.3&68.5&49.0&69.6
            &96.4&88.7&75.2&85.0
            &27.0&17.8&45.3&30.8
            &14.60&9.89&6.61& 7.98\\
            DINOv2-ViT-B/14
            &81.2&68.4&51.5&69.4
            &94.8&87.8&76.8&87.8
            &29.1&18.1&42.1&30.9
            & 7.40 &9.98&15.60 & 10.46 \\
            Stable Diffusion
            &92.1&62.6&48.4&72.5
            &97.5&80.1&64.7&83.8
            &22.1&26.5&69.1&37.5
            & 3.49 & 8.98 &9.31 & 6.41\\
            \midrule
            \multicolumn{17}{c}{DINOv1-ViT-S/8 + DINOv2-ViT-B/14} \\
            \midrule
            Concat
            &83.5&64.7&52.0&69.9
            &96.4&84.8&77.3&88.1
            &26.6&23.1&43.2&31.0
            &6.23&13.50&14.60&10.31\\
            FMap
            &84.9&70.5&53.6&72.2
            &96.6&89.1&81.2&90.3
            &25.6&18.2&37.2&27.7
            &5.26&9.61&11.10&7.95\\
            \midrule
            \multicolumn{17}{c}{DINOv2-ViT-S/14 + Stable Diffusion} \\
            \midrule
            Concat
            &93.3&72.7&57.7&78.1
            &98.5&89.2&77.1&89.9
            &20.0&17.4&46.0&27.5
            &3.91&7.65&10.10&6.58\\
            FMap
            &84.7&70.2&51.8&71.5
            &98.6&88.8&77.2&90.0
            &18.7&19.2&43.2&26.3
            &3.77&7.85&9.75&6.47\\
            \midrule
            \multicolumn{17}{c}{DINOv2-ViT-B/14 + Stable Diffusion} \\
            \midrule
            Concat
            &92.9&73.8&59.6&78.7
            &98.5&90.1&78.9&90.7
            &20.1&16.7&42.8&26.4
            & 4.12 & 7.75 & 10.42 & 6.81\\
            FMap
            &93.1&74.0&59.9&78.9
            &97.9&89.5&81.5&91.1
            &24.3&17.1&34.9&26.1
            & 3.15 & 8.29 & 7.70 & 5.74\\
            \midrule
            \multicolumn{17}{c}{DINOv2-ViT-S/14 layer9 + layer11} \\
            \midrule
            Layer9
            &82.4&63.2&45.9&67.2
            &96.0&83.3&68.3&84.8
            &28.1&22.1&59.3&36.5
            &6.62&10.21&14.02&9.64\\
            Layer11
            &83.3&68.5&51.5&69.6
            &96.4&88.7&76.8&85.0
            &27.0&17.8&42.1&30.8
            &6.61&9.59&15.60&7.98\\
            Concat
            &74.6&67.3&51.0&70.5
            &86.8&86.9&75.4&88.1
            &59.8&18.9&46.8&31.0
            &12.53&9.65&13.66&9.25\\
            FMap
            &84.2&68.1&51.6&70.8
            &96.9&87.4&77.8&89.1
            &25.6&18.2&41.9&29.1
            &5.03&8.15&8.06&6.60\\
            \midrule
            \multicolumn{17}{c}{DINOv2-ViT-B/14 layer9 + layer11} \\
            \midrule
            Layer9
            &60.4&64.2&47.9&57.2
            &94.7&84.1&71.8&85.4
            &29.9&21.2&50.4&34.5
            &7.38&10.70&15.85&10.66\\
            Layer11
            &81.2&68.4&51.5&69.4
            &94.8&87.8&76.8&87.8
            &29.1&18.1&42.1&30.9
            & 7.40 &9.98&15.60 & 10.46\\
            Concat
            &82.5&67.1&52.4&70.0
            &95.6&86.7&76.7&87.9
            &28.1&18.8&43.5&30.9
            &6.95&10.10&15.02&10.24\\
            FMap
            &83.1&67.2&53.4&70.6
            &95.7&87.7&82.1&89.8
            &22.8&18.1&35.9&25.9
            &5.95&8.60&11.90&8.27\\
            \bottomrule
        \end{tabular}
    }
\end{table}

\vspace{-3em}

\begin{table*}[ht!]
    \centering
    \footnotesize
    \caption{\textbf{Per-category results on SPair-71k (\cref{tab:exp:spair}, \cref{sec:exp:spair}).
    }}
    \vspace{-.5em}
    \setlength{\tabcolsep}{2pt}
    \resizebox{\linewidth}{!}{%
        \begin{tabular}{lxxxxxxxxxxxxxxxxxxy}
            \toprule
            \textbf{Method} & \textbf{Aero} & \textbf{Bike} & \textbf{Bird} & \textbf{Boat} & \textbf{Bottle} & \textbf{Bus} & \textbf{Car} & \textbf{Cat} & \textbf{Chair} & \textbf{Cow} & \textbf{Dog} & \textbf{Horse} & \textbf{Motor} & \textbf{Person} & \textbf{Plant} & \textbf{Sheep} & \textbf{Train} & \textbf{TV}& \textbf{Avg.} \\
            \midrule
            \multicolumn{20}{c}{PCK@0.10$\uparrow$}\\
            \midrule
            DINOv2 &67.7&63.3&83.8&39.1&45.6&44.0&41.0&68.3&33.9&67.5&48.8&61.2&64.1&61.3&22.2&60.9&46.1&23.1&52.3 \\
            SD &56.5&55.0&75.0&32.2&51.9&45.3&34.3&73.4&34.8&70.6&46.3&63.7&44.7&49.6&50.8&51.3&55.7&32.1&51.2 \\
            Concat &67.7&60.8&83.8&40.2&52.2&50.7&37.3&78.9&42.5&68.9&53.9&61.3&61.2&64.6&42.9&57.4&60.0&46.4 & 57.2\\
            FMap & 65.3&65.0&83.8&39.1&47.5&44.0&40.3&69.4&37.4&64.4&59.2&63.1&66.0&64.7&27.0&60.0&51.3&47.4 & 55.3\\
            \midrule
            \multicolumn{20}{c}{PCK@0.20$\uparrow$}\\
            \midrule
            DINOv2 & 77.7&78.3&91.9&56.3&63.9&54.7&48.5&85.4&50.4&84.1&63.4&86.0&84.5&71.4&47.6&75.7&67.8&36.3&68.0\\
            SD & 70.5&73.3&90.4&41.4&65.2&52.6&41.0&81.4&47.8&78.5&60.3&78.3&55.3&62.2&69.8&64.3&72.6&50.0& 64.1\\
            Concat & 78.2&81.7&94.1&56.3&69.6&60.7&44.0&89.9&56.5&80.4&71.9&80.3&77.6&76.5&65.9&71.3&78.7&67.5& 72.2\\
            FMap & 81.2&80.0&94.9&56.3&67.1&57.3&56.0&82.9&47.8&78.5&79.9&80.3&82.5&80.7&57.9&75.7&73.0&68.0 & 72.6\\
            \midrule
            \multicolumn{20}{c}{MSE$\downarrow$}\\
            \midrule
            DINOv2 &101.5&58.7&40.2&162.6&131.7&160.9&141.7&58.3&120.6&60.7&72.0&76.4&62.8&100.6&167.5&65.5&82.6&225.9&105.0\\
            SD &  122.3&77.7&57.7&225.4&132.8&178.1&158.2&70.6&128.5&74.2&109.2&86.5&144.1&111.4&118.3&110.4&77.9&186.5&120.5\\
            Concat &98.2&64.9&39.9&155.3&110.7&156.9&150.4&47.2&105.6&65.5&78.8&94.7&85.7&84.1&143.1&94.1&64.4&111.3&97.2\\
            FMap &88.9&54.8&37.7&162.9&116.3&153.8&123.1&57.7&93.1&70.3&60.6&96.9&66.0&74.9&114.7&68.8&73.7&103.1&88.0\\
            \bottomrule
        \end{tabular}%
    }%
\end{table*}

%% file: main.bbl
\begin{thebibliography}{10}
\providecommand{\url}[1]{\texttt{#1}}
\providecommand{\urlprefix}{URL }
\providecommand{\doi}[1]{https://doi.org/#1}

\bibitem{amir2021deep}
Amir, S., Gandelsman, Y., Bagon, S., Dekel, T.: Deep vit features as dense visual descriptors. arXiv preprint arXiv:2112.05814  \textbf{2}(3), ~4 (2021)

\bibitem{attaiki2021dpfm}
Attaiki, S., Pai, G., Ovsjanikov, M.: Dpfm: Deep partial functional maps. In: International Conference on 3D Vision (3DV) (2021)

\bibitem{aubry2011wave}
Aubry, M., Schlickewei, U., Cremers, D.: The wave kernel signature: A quantum mechanical approach to shape analysis. In: ICCV Workshops (2011)

\bibitem{burghard2017embedding}
Burghard, O., Dieckmann, A., Klein, R.: Embedding shapes with green’s functions for global shape matching. Computers \& Graphics  \textbf{68},  1--10 (2017)

\bibitem{cao2022unsupervised}
Cao, D., Bernard, F.: Unsupervised deep multi-shape matching. In: European Conference on Computer Vision (ECCV) (2022)

\bibitem{caron2021emerging}
Caron, M., Touvron, H., Misra, I., J{\'e}gou, H., Mairal, J., Bojanowski, P., Joulin, A.: Emerging properties in self-supervised vision transformers. In: International Conference on Computer Vision (ICCV) (2021)

\bibitem{cho2021cats}
Cho, S., Hong, S., Jeon, S., Lee, Y., Sohn, K., Kim, S.: Cats: Cost aggregation transformers for visual correspondence. Advances in Neural Information Processing Systems  \textbf{34},  9011--9023 (2021)

\bibitem{donati2022deep}
Donati, N., Corman, E., Ovsjanikov, M.: Deep orientation-aware functional maps: Tackling symmetry issues in shape matching. In: Conference on Computer Vision and Pattern Recognition (CVPR) (2022)

\bibitem{dusmanu2019d2}
Dusmanu, M., Rocco, I., Pajdla, T., Pollefeys, M., Sivic, J., Torii, A., Sattler, T.: D2-net: A trainable cnn for joint description and detection of local features. In: Conference on Computer Vision and Pattern Recognition (CVPR) (2019)

\bibitem{gupta2023asic}
Gupta, K., Jampani, V., Esteves, C., Shrivastava, A., Makadia, A., Snavely, N., Kar, A.: Asic: Aligning sparse in-the-wild image collections. arXiv preprint arXiv:2303.16201  (2023)

\bibitem{halimi2019unsupervised}
Halimi, O., Litany, O., Rodola, E., Bronstein, A.M., Kimmel, R.: Unsupervised learning of dense shape correspondence. In: Conference on Computer Vision and Pattern Recognition (CVPR) (2019)

\bibitem{hariharan2011semantic}
Hariharan, B., Arbel{\'a}ez, P., Bourdev, L., Maji, S., Malik, J.: Semantic contours from inverse detectors. In: International Conference on Computer Vision (ICCV) (2011)

\bibitem{hedlin2023unsupervised}
Hedlin, E., Sharma, G., Mahajan, S., Isack, H., Kar, A., Tagliasacchi, A., Yi, K.M.: Unsupervised semantic correspondence using stable diffusion. arXiv preprint arXiv:2305.15581  (2023)

\bibitem{huang2014functional}
Huang, Q., Wang, F., Guibas, L.: Functional map networks for analyzing and exploring large shape collections. ACM Transactions on Graphics (TOG)  \textbf{33}(4),  1--11 (2014)

\bibitem{jacobs1991adaptive}
Jacobs, R.A., Jordan, M.I., Nowlan, S.J., Hinton, G.E.: Adaptive mixtures of local experts. Neural computation  \textbf{3}(1),  79--87 (1991)

\bibitem{jeon2018parn}
Jeon, S., Kim, S., Min, D., Sohn, K.: Parn: Pyramidal affine regression networks for dense semantic correspondence. In: European Conference on Computer Vision (ECCV) (2018)

\bibitem{kim2018recurrent}
Kim, S., Lin, S., Jeon, S.R., Min, D., Sohn, K.: Recurrent transformer networks for semantic correspondence. In: Advances in Neural Information Processing Systems (NeurIPS) (2018)

\bibitem{kovnatsky2013coupled}
Kovnatsky, A., Bronstein, M.M., Bronstein, A.M., Glashoff, K., Kimmel, R.: Coupled quasi-harmonic bases. In: Computer Graphics Forum (2013)

\bibitem{learned2005data}
Learned-Miller, E.G.: Data driven image models through continuous joint alignment. Transactions on Pattern Analysis and Machine Intelligence (TPAMI)  \textbf{28}(2),  236--250 (2005)

\bibitem{li2022learning}
Li, L., Donati, N., Ovsjanikov, M.: Learning multi-resolution functional maps with spectral attention for robust shape matching. In: Advances in Neural Information Processing Systems (NeurIPS) (2022)

\bibitem{lin2014jointly}
Lin, Y.L., Morariu, V.I., Hsu, W., Davis, L.S.: Jointly optimizing 3d model fitting and fine-grained classification. In: European Conference on Computer Vision (ECCV) (2014)

\bibitem{litany2017deep}
Litany, O., Remez, T., Rodola, E., Bronstein, A., Bronstein, M.: Deep functional maps: Structured prediction for dense shape correspondence. In: International Conference on Computer Vision (ICCV) (2017)

\bibitem{liu2010sift}
Liu, C., Yuen, J., Torralba, A.: Sift flow: Dense correspondence across scenes and its applications. Transactions on Pattern Analysis and Machine Intelligence (TPAMI)  \textbf{33}(5),  978--994 (2010)

\bibitem{liu2020semantic}
Liu, Y., Zhu, L., Yamada, M., Yang, Y.: Semantic correspondence as an optimal transport problem. In: Proceedings of the IEEE/CVF Conference on Computer Vision and Pattern Recognition. pp. 4463--4472 (2020)

\bibitem{min2019spair}
Min, J., Lee, J., Ponce, J., Cho, M.: Spair-71k: A large-scale benchmark for semantic correspondence. arXiv preprint arXiv:1908.10543  (2019)

\bibitem{myers2015affordance}
Myers, A., Teo, C.L., Ferm{\"u}ller, C., Aloimonos, Y.: Affordance detection of tool parts from geometric features. In: International Conference on Robotics and Automation (ICRA) (2015)

\bibitem{nogneng2017informative}
Nogneng, D., Ovsjanikov, M.: Informative descriptor preservation via commutativity for shape matching. In: Computer Graphics Forum (2017)

\bibitem{ofri2023neural}
Ofri-Amar, D., Geyer, M., Kasten, Y., Dekel, T.: Neural congealing: Aligning images to a joint semantic atlas. In: Conference on Computer Vision and Pattern Recognition (CVPR) (2023)

\bibitem{ono2018lf}
Ono, Y., Trulls, E., Fua, P., Yi, K.M.: Lf-net: Learning local features from images. In: Advances in Neural Information Processing Systems (NeurIPS) (2018)

\bibitem{oquab2023dinov2}
Oquab, M., Darcet, T., Moutakanni, T., Vo, H., Szafraniec, M., Khalidov, V., Fernandez, P., Haziza, D., Massa, F., El-Nouby, A., et~al.: Dinov2: Learning robust visual features without supervision. arXiv preprint arXiv:2304.07193  (2023)

\bibitem{ovsjanikov2012functional}
Ovsjanikov, M., Ben-Chen, M., Solomon, J., Butscher, A., Guibas, L.: Functional maps: a flexible representation of maps between shapes. ACM Transactions on Graphics (TOG)  \textbf{31}(4),  1--11 (2012)

\bibitem{peebles2022gan}
Peebles, W., Zhu, J.Y., Zhang, R., Torralba, A., Efros, A.A., Shechtman, E.: Gan-supervised dense visual alignment. In: Conference on Computer Vision and Pattern Recognition (CVPR) (2022)

\bibitem{revaud2019r2d2}
Revaud, J., De~Souza, C., Humenberger, M., Weinzaepfel, P.: R2d2: Reliable and repeatable detector and descriptor. In: Advances in Neural Information Processing Systems (NeurIPS) (2019)

\bibitem{rocco2017convolutional}
Rocco, I., Arandjelovic, R., Sivic, J.: Convolutional neural network architecture for geometric matching. In: Conference on Computer Vision and Pattern Recognition (CVPR) (2017)

\bibitem{rocco2018end}
Rocco, I., Arandjelovi{\'c}, R., Sivic, J.: End-to-end weakly-supervised semantic alignment. In: Conference on Computer Vision and Pattern Recognition (CVPR) (2018)

\bibitem{rodola2017partial}
Rodol{\`a}, E., Cosmo, L., Bronstein, M.M., Torsello, A., Cremers, D.: Partial functional correspondence. In: Computer Graphics Forum (2017)

\bibitem{rombach2022high}
Rombach, R., Blattmann, A., Lorenz, D., Esser, P., Ommer, B.: High-resolution image synthesis with latent diffusion models. In: Conference on Computer Vision and Pattern Recognition (CVPR) (2022)

\bibitem{roufosse2019unsupervised}
Roufosse, J.M., Sharma, A., Ovsjanikov, M.: Unsupervised deep learning for structured shape matching. In: International Conference on Computer Vision (ICCV) (2019)

\bibitem{rubinstein2013unsupervised}
Rubinstein, M., Joulin, A., Kopf, J., Liu, C.: Unsupervised joint object discovery and segmentation in internet images. In: Conference on Computer Vision and Pattern Recognition (CVPR) (2013)

\bibitem{sarlin2020superglue}
Sarlin, P.E., DeTone, D., Malisiewicz, T., Rabinovich, A.: Superglue: Learning feature matching with graph neural networks. In: Conference on Computer Vision and Pattern Recognition (CVPR) (2020)

\bibitem{seo2018attentive}
Seo, P.H., Lee, J., Jung, D., Han, B., Cho, M.: Attentive semantic alignment with offset-aware correlation kernels. In: European Conference on Computer Vision (ECCV) (2018)

\bibitem{sharp2022diffusionnet}
Sharp, N., Attaiki, S., Crane, K., Ovsjanikov, M.: Diffusionnet: Discretization agnostic learning on surfaces. ACM Transactions on Graphics (TOG)  \textbf{41}(3),  1--16 (2022)

\bibitem{sun2009concise}
Sun, J., Ovsjanikov, M., Guibas, L.: A concise and provably informative multi-scale signature based on heat diffusion. In: Computer Graphics Forum (2009)

\bibitem{tang2023emergent}
Tang, L., Jia, M., Wang, Q., Phoo, C.P., Hariharan, B.: Emergent correspondence from image diffusion. arXiv preprint arXiv:2306.03881  (2023)

\bibitem{taniai2016joint}
Taniai, T., Sinha, S.N., Sato, Y.: Joint recovery of dense correspondence and cosegmentation in two images. In: Conference on Computer Vision and Pattern Recognition (CVPR) (2016)

\bibitem{truong2020gocor}
Truong, P., Danelljan, M., Gool, L.V., Timofte, R.: Gocor: Bringing globally optimized correspondence volumes into your neural network. In: Advances in Neural Information Processing Systems (NeurIPS) (2020)

\bibitem{truong2020glu}
Truong, P., Danelljan, M., Timofte, R.: Glu-net: Global-local universal network for dense flow and correspondences. In: Conference on Computer Vision and Pattern Recognition (CVPR) (2020)

\bibitem{truong2021learning}
Truong, P., Danelljan, M., Van~Gool, L., Timofte, R.: Learning accurate dense correspondences and when to trust them. In: Conference on Computer Vision and Pattern Recognition (CVPR) (2021)

\bibitem{truong2021warp}
Truong, P., Danelljan, M., Yu, F., Van~Gool, L.: Warp consistency for unsupervised learning of dense correspondences. In: International Conference on Computer Vision (ICCV) (2021)

\bibitem{truong2022probabilistic}
Truong, P., Danelljan, M., Yu, F., Van~Gool, L.: Probabilistic warp consistency for weakly-supervised semantic correspondences. In: Conference on Computer Vision and Pattern Recognition (CVPR) (2022)

\bibitem{tyszkiewicz2020disk}
Tyszkiewicz, M., Fua, P., Trulls, E.: Disk: Learning local features with policy gradient. In: Advances in Neural Information Processing Systems (NeurIPS) (2020)

\bibitem{wang2013image}
Wang, F., Huang, Q., Guibas, L.J.: Image co-segmentation via consistent functional maps. In: International Conference on Computer Vision (ICCV) (2013)

\bibitem{wang2014unsupervised}
Wang, F., Huang, Q., Ovsjanikov, M., Guibas, L.J.: Unsupervised multi-class joint image segmentation. In: Conference on Computer Vision and Pattern Recognition (CVPR) (2014)

\bibitem{yang2012articulated}
Yang, Y., Ramanan, D.: Articulated human detection with flexible mixtures of parts. Transactions on Pattern Analysis and Machine Intelligence (TPAMI)  \textbf{35}(12),  2878--2890 (2012)

\bibitem{yi2016lift}
Yi, K.M., Trulls, E., Lepetit, V., Fua, P.: Lift: Learned invariant feature transform. In: European Conference on Computer Vision (ECCV) (2016)

\bibitem{zhang2023tale}
Zhang, J., Herrmann, C., Hur, J., Cabrera, L.P., Jampani, V., Sun, D., Yang, M.H.: A tale of two features: Stable diffusion complements dino for zero-shot semantic correspondence. arXiv preprint arXiv:2305.15347  (2023)

\end{thebibliography}
